\pdfoutput=1

\documentclass[11pt]{article}

\usepackage[final]{ACL2023}

\usepackage{times}
\usepackage{latexsym}
\usepackage{pifont}
\usepackage{newunicodechar}
\usepackage{booktabs}
\usepackage{algorithm}
\usepackage{multirow}
\usepackage{array}
\usepackage{subcaption}
\usepackage{bbm}
\usepackage{enumitem}
\usepackage[noend]{algpseudocode}
\newunicodechar{✓}{\ding{51}}
\newunicodechar{✗}{\ding{55}}

\usepackage[T1]{fontenc}

\usepackage[utf8]{inputenc}
\usepackage{xspace}

\newcommand{\ours}{\textsc{DISCO}\xspace}
\usepackage{microtype}
\usepackage{amsmath}
\usepackage{graphicx}
\usepackage{amssymb}

\usepackage{inconsolata}
\usepackage[normalem]{ulem}
\newcommand{\mirrorball}{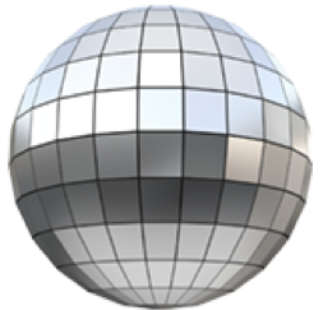}
\usepackage[export]{adjustbox}
\newcommand{\MyEmoji}[1]{\includegraphics[width=1em,valign=t]{#1}}
\useunder{\uline}{\ul}{}

\title{\MyEmoji{\mirrorball} DISCO Balances the Scales: Adaptive Domain- and Difficulty-Aware Reinforcement Learning on Imbalanced Data}

\author{Yuhang Zhou$^{\ 1*}$\quad Jing Zhu$^{\ 3*}$ \quad Shengyi Qian$^{\ 3}$ \quad Zhuokai Zhao$^{\ 4}$ \quad Xiyao Wang$^{\ 1}$ \quad  \\ \textbf{Xiaoyu Liu}$^{\ 1}$ \quad \textbf{Ming Li}$^{\ 1}$\quad \textbf{Paiheng Xu}$^{\ 1}$ \quad \textbf{Wei Ai}$^{\ 1}$ \quad \textbf{Furong Huang}$^{\ 1, 2}$\\
         $^{1}$ University of Maryland, College Park \quad $^{2}$ Capital One \\ $^{3}$ University of Michigan, Ann Arbor \quad $^{4}$ University of Chicago \\ \texttt{\{tonyzhou, xywang, xliu1231, paiheng, minglii, aiwei, furongh\}@umd.edu} \\
         \texttt{\{jingzhuu, syqian\}@umich.edu} \quad \texttt{zhuokai@uchicago.edu}}

\begin{document}
\maketitle
\begin{abstract}

Large Language Models (LLMs) are increasingly aligned with human preferences through Reinforcement Learning from Human Feedback (RLHF). Among RLHF methods, Group Relative Policy Optimization (GRPO) has gained attention for its simplicity and strong performance, notably eliminating the need for a learned value function. However, GRPO implicitly assumes a balanced domain distribution and uniform semantic alignment across groups---assumptions that rarely hold in real-world datasets. When applied to multi-domain, imbalanced data, GRPO disproportionately optimizes for dominant domains, neglecting underrepresented ones and resulting in poor generalization and fairness.
We propose \textbf{D}omain-\textbf{I}nformed \textbf{S}elf-\textbf{C}onsistency Policy \textbf{O}ptimization (\textbf{\ours}), a principled extension to GRPO that addresses inter-group imbalance with two key innovations. \emph{Domain-aware reward scaling} counteracts frequency bias by reweighting optimization based on domain prevalence. \emph{Difficulty-aware reward scaling} leverages prompt-level self-consistency to identify and prioritize uncertain prompts that offer greater learning value. Together, these strategies promote more equitable and effective policy learning across domains.
Extensive experiments across multiple LLMs and skewed training distributions show that \ours improves generalization, outperforms existing GRPO variants by 5\% on Qwen3 models, and sets new state-of-the-art results on multi-domain alignment benchmarks. Our code and data are available at \url{https://github.com/Tonyzhou98/disco_grpo}

%
\end{abstract}

\let\thefootnote\relax\footnote{*Equal contribution.}
\setcounter{footnote}{0}

\section{Introduction}
\label{sec:intro}


Aligning large language models (LLMs) with human preferences is a central challenge in modern AI systems~\cite{openai2023gpt4, touvron2023llama, qwen2.5}. Reinforcement Learning from Human Feedback (RLHF) has become the dominant approach for fine-tuning LLMs toward desirable behavior, enabling alignment with nuanced human intent~\cite{ouyang2022training, shao2024deepseekmath, zheng2023judging, rafailov2023direct, zhou2023lima, wang2024preference, wang2025beyond}. Within this framework, Group Relative Policy Optimization (GRPO)~\cite{shao2024deepseekmath} offers a promising alternative to value-based methods, simplifying training while achieving strong performance.

\begin{figure*}[tbp]
  \centering
  \includegraphics[width=\linewidth]{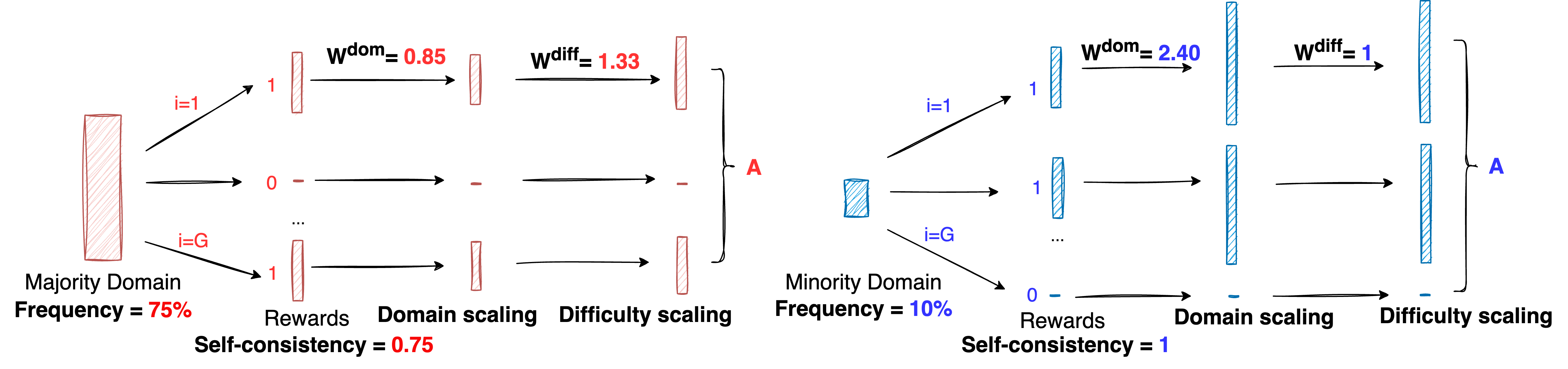}
  \vspace{-0.3cm}
   \caption{\label{fig:overview} \textbf{Overview of the proposed \ours scaling framework.} The framework is composed of two strategies to enhance GRPO's robustness: (1) domain-aware scaling, which reweights prompt groups based on domain frequency, and (2) difficulty-aware scaling, which encourages the model to focus more on uncertain samples based on self-consistency scores. $w^{\text{dom}}$ and $w^{\text{diff}}$ denote the domain and difficulty weight. \underline{G}=Group size, \underline{A}=Advantage.} 
   \vspace{-0.3cm}
\end{figure*}

Despite its advantages, GRPO faces a significant challenge when applied to multi-domain datasets, a common scenario in real-world applications spanning math, question answering, and commonsense reasoning. While GRPO effectively removes the need for a value network and mitigates within-group variance, it implicitly assumes that prompt groups are sampled uniformly and that reward signals are semantically aligned across domains. However, this assumption often fails in practice. Real-world datasets are typically imbalanced, with a few dominant domains and many underrepresented ones~\cite{zhou2024multi}. Optimization gradients become skewed toward high-frequency domains, starving rare domains of learning signal. This results in models that generalize poorly to critical low-resource domains and amplify existing data biases. Data augmentation offers one workaround but introduces substantial overhead in generating high-quality synthetic prompts~\cite{tepper2020balancing}.


To address the inter-group imbalance in GRPO, we propose \textbf{D}omain-\textbf{I}nformed \textbf{S}elf-\textbf{C}onsistency Policy \textbf{O}ptimization (\textbf{\ours{}}), an enhanced framework designed to promote equitable learning across imbalanced multi-domain datasets. \ours introduces two complementary strategies to improve generalization under distributional skew: \emph{domain-aware} and \emph{difficulty-aware} reward scaling.

\textit{Domain-aware scaling} reweights prompt groups inversely by their frequency, reducing over-optimization on dominant domains while amplifying learning signals from underrepresented ones.
\textit{Difficulty-aware scaling} leverages prompt-level self-consistency, an intrinsic signal in GRPO, to identify and upweight prompts where the policy exhibits high uncertainty or inconsistent responses. Since not all prompts are equally challenging, treating them uniformly can lead the policy to overfit on easy examples while neglecting harder, more informative ones. By prioritizing uncertain prompts, this strategy guides the model to focus its learning on cases that offer a greater signal for improvement.

By integrating these two forms of adaptive scaling, \ours enables all domains, regardless of their prevalence, to meaningfully contribute to policy optimization. As a result, it effectively mitigates GRPO's inter-group imbalance and achieves state-of-the-art performance across diverse LLM architectures and training distributions.

Our contributions are summarized as follows: 

\begin{itemize}[leftmargin=*]
    \item \textbf{Systematic Analysis of GRPO under data imbalance:}  We perform the first systematic analysis on GRPO's inherent vulnerability to dataset imbalance, empirically demonstrating its resulting performance degradation on underrepresented domains.

    \item \textbf{Strategic Framework:} Our novel \ours framework introduces a powerful integration of domain-aware and difficulty-aware reward scaling strategies, effectively addressing the inter-group imbalance that limits standard GRPO.
    
    \item \textbf{SoTA Performance:} 
    Our comprehensive empirical evaluations confirm that \ours improves existing GRPO algorithms by 5\% on Qwen3 models across diverse benchmarks and sets new standards for generalization performance.

\end{itemize}

\section{Background and Motivation}

This section first provides the necessary background on the GRPO algorithm. We then present an empirical study to concretely demonstrate how its optimization behavior is affected by domain imbalance in training data, thereby establishing the specific motivation for the enhancements proposed in subsequent sections.

\subsection{Group Relative Policy Optimization}

While standard RLHF relies on learned value networks, GRPO~\cite{shao2024deepseekmath} computes relative advantages directly within each prompt group, simplifying training. Specifically, for a given prompt $q$, $G$ candidate outputs $\{o_i\}_{i=1}^G$ are sampled from the old policy model $\pi_{\theta_{\text{old}}}$. A reward $r_i$ is assigned to each $o_i$, and the group-normalized advantage $A_i$ is calculated by normalizing the rewards across the group, which is:
\begin{equation}
\label{eq:ori_adv}
A_i = \frac{r_i - \bar{r}}{\sigma_r}
\end{equation}
where $\bar{r} = \frac{1}{G} \sum_{j=1}^{G} r_j$ is the group mean reward, and $\sigma_r$ is the standard deviation of the rewards within the group. 
The overall GRPO objective is given by:

\vspace{-0.5cm}
\scalebox{0.8}{%
\begin{minipage}{\linewidth}
\centering
\begin{multline} 
\label{eq:grpo_objective_multline}
\mathcal{J}_{\text{GRPO}}(\theta) = \mathbb{E}_{q, \{o_i\}} \Biggl[ \frac{1}{G} \sum_{i=1}^{G} \min \Biggl( \frac{\pi_\theta(o_i \mid q)}{\pi_{\theta_{\text{old}}}(o_i \mid q)} A_i, \\
   \text{clip}\left(\frac{\pi_\theta(o_i \mid q)}{\pi_{\theta_{\text{old}}}(o_i \mid q)}, 1-\epsilon, 1+\epsilon\right) A_i \Biggr) \Biggr]  - \beta \, \mathbb{D}_{\text{KL}}[\pi_\theta \| \pi_{\text{ref}}]
\end{multline}
\end{minipage}
}

\noindent where $\pi_\theta$ and $\pi_{\theta_{\text{old}}}$ denote the new and old policies respectively, $A_i$ is the group-normalized advantage, `clip' denotes the clip function, and $\mathbb{D}_{\text{KL}}$ is a KL regularization term encouraging the updated policy to remain close to a reference policy.

In this work, we focus on a rule-based GRPO setup, similar to DeepSeek's R1 model training~\cite{guo2025deepseek}, with deterministic rewards: 1 for an exact match (EM) with the ground-truth answer, 0 otherwise. This controlled setup allows us to isolate optimization and data distribution effects on learning dynamics, avoiding the variability introduced by learned reward models.

\subsection{Empirical Demonstration of Domain Imbalance Impact}
\label{sec:imbalance_results}

\begin{table}[tbp]
\resizebox{\linewidth}{!}{%
\centering
\begin{tabular}{lccccc|c}
\toprule
&           IMDB &          GSM8K &           Math &             NQ &            ARC &            Avg. \\
\midrule
      Balanced &          89.60 &          57.94 &          23.82 &          14.90 &          47.44 & \textbf{46.74} \\
    Math heavy &          87.80 & \textbf{58.98} & \textbf{24.38} &          13.96 &          45.99 &          46.22 \\
    IMDB heavy & \textbf{90.90} &          56.88 &          23.54 &          13.85 &          45.90 &          46.21 \\
      NQ heavy &          89.40 &          56.22 &          23.36 & \textbf{16.57} &          46.58 &          46.43 \\
     ARC heavy &          88.80 &          55.29 &          22.46 &          14.16 & \textbf{48.29} &          45.80 \\
\bottomrule
\end{tabular}
}
\caption{Performance (Exact Match (EM) Accuracy \%) of Qwen2.5-0.5B trained with GRPO ($G=4$) under various domain-heavy training distributions.}
\label{tab:grpo_imbalance}
\end{table}

While GRPO normalizes advantages locally within prompt groups, its global optimization trajectory can be unduly influenced by the frequency of domains in the training data, a vulnerability already highlighted in Section~\ref{sec:intro}. To empirically demonstrate this limitation, we performed an experiment training Qwen2.5-0.5B~\cite{qwen2.5} with GRPO using datasets featuring distinct domain compositions and present the performance in Table \ref{tab:grpo_imbalance}. Specifically, we trained multiple models, each on 4,000 examples. For each model, the training data composition was intentionally skewed by heavily weighting one domain (e.g., 3,000 math prompts for a `Math-heavy' model) while underrepresenting the others. Table~\ref{tab:data_overview} and Appendix~\ref{sec:implement} provide details on data and implementation.

As shown in Table~\ref{tab:grpo_imbalance}, domain-heavy training biases GRPO performance: models excel in the overrepresented domain but underperform on others compared to a balanced setup. For example, the Math-heavy model performs best on math tasks (24.38\% Math, 58.98\% GSM8K), but its average score across all domains is lower than the balanced model's, primarily due to sub-optimal performance on other domains such as IMDB and ARC. This trade-off is consistent across other skewed settings, where specialization gains are accompanied by performance drops in underrepresented domains.

These results confirm that GRPO lacks a mechanism for inter-group calibration, and its optimization may become biased toward more frequently sampled domains, leading to suboptimal performance on rare but important tasks. This observation motivates our investigation into domain-aware and difficulty-aware reward scaling methods to improve robustness in imbalanced settings.

\section{Domain- and Difficulty-Aware Scaling}

Building on the observation that GRPO lacks mechanisms for addressing domain-level imbalance, our \ours, as depicted in Figure \ref{fig:overview}, proposes two modifications in multi-domain settings: \textbf{domain-aware} and \textbf{difficulty-aware reward scaling}. 

First, we introduce domain-aware scaling, which reweights prompt groups based on domain frequency. This strengthens the learning signal for underrepresented ones.

Secondly, our difficulty-aware scaling enhances learning efficiency by focusing on challenging examples. It uses prompt-level self-consistency, the average reward from $G$ candidate outputs per prompt, as a signal of policy uncertainty within GRPO. A lower self-consistency indicates higher uncertainty and thus greater difficulty. Group rewards are then scaled by a weight inversely proportional to this self-consistency score, thereby prioritizing more ambiguous prompts.

Together, these two mechanisms allow GRPO to adaptively rescale group-level advantages based on both domain rarity and prompt difficulty, offering a more globally informed optimization process while preserving its core group-relative framework.

\begin{table}[tbp]
\centering
\begin{tabular}{c}
\begin{subtable}[t]{0.93\linewidth}
\centering
\setlength{\tabcolsep}{2pt}
\renewcommand{\arraystretch}{1}
\resizebox{\linewidth}{!}{%
\begin{tabular}{ll}
\toprule
\textbf{Dataset} & \textbf{Task Domain} \\
\midrule
IMDB \cite{maas2011learning} & Text Classification (TC) \\
GSM8K \cite{cobbe2021training} & Mathematical Reasoning \\
MATH \cite{hendrycks2021measuring} & Mathematical Reasoning \\
Natural Questions (NQ) \cite{lee2019latent} & Open-domain QA \\
ARC \cite{allenai:arc} & Multi-step Reasoning QA \\
\bottomrule
\end{tabular}
}
\caption{Test datasets and their corresponding domains. ARC denotes both the ARC-Easy and -Challenge subsets. All evaluations are conducted using exact match (EM).}
\end{subtable}
\\[1.5ex]
\begin{subtable}[t]{0.93\linewidth}
\centering
\setlength{\tabcolsep}{12pt}
\renewcommand{\arraystretch}{1}
\resizebox{\linewidth}{!}{%
\begin{tabular}{lcccc}
\toprule
\textbf{Setup} & \textbf{Math} & \textbf{NQ} & \textbf{ARC} & \textbf{IMDB} \\
\midrule
Balanced     & 25\% & 25\% & 25\% & 25\% \\
Math-heavy   & 75\% & 8.3\% & 8.3\% & 8.4\% \\
NQ-heavy     & 8.3\% & 75\% & 8.3\% & 8.4\% \\
ARC-heavy    & 8.3\% & 8.3\% & 75\% & 8.4\% \\
IMDB-heavy   & 8.3\% & 8.3\% & 8.4\% & 75\% \\
\bottomrule
\end{tabular}
}
\caption{Training Distribution by Domain (Proportions)}
\end{subtable}
\end{tabular}
\caption{Overview of evaluation and training datasets. ``Heavy'' settings allocate 75\% of training prompts to a single domain, with the remaining 25\% distributed equally among the others. For the math domain, we sample from the MetaMath dataset~\cite{yu2023metamath} for training, while for all other domains, we use the training portion of each corresponding evaluation benchmark.}
\label{tab:data_overview}
\end{table}

\subsection{Domain-Aware Scaling}
\label{sec:method_domain}
Vanilla GRPO's group-level normalization operates independently of a prompt group's originating domain frequency. In imbalanced datasets (Section \ref{sec:imbalance_results}), this means high-frequency domains disproportionately influence the aggregated optimization gradient, potentially marginalizing low-frequency domains. To mitigate this issue, we introduce a \textbf{domain-aware reward scaling} strategy. For each prompt group $q$ from domain $d$, we apply a domain weight $w^{\text{dom}}(q)$ to rescale its rewards:
\begin{equation}
r_i^{\text{scaled}} = r_i \cdot w^{\text{dom}}(q),
\end{equation}
and compute the group-level advantage as:
\begin{equation}
\label{eq:scale_adv}
A_i = r_i^{\text{scaled}} - \bar{r}^{\text{scaled}},
\end{equation}
where $\bar{r}^{\text{scaled}} = \frac{1}{G} \sum_{j=1}^G r_j^{\text{scaled}}$ is the group mean. 

Note that we do not apply standard deviation normalization, as in the original GRPO (Equation~\ref{eq:ori_adv}), in order to preserve the absolute scaling effect of the domain weights. This design allows domain frequency to directly modulate the magnitude of the advantage signal, enabling rarer domains to exert a stronger influence during policy updates.

\paragraph{Domain Weight Variants.}
We explore three definitions of the domain weight based on domain frequency. Let $p_d$ denote the proportion of prompt groups from domain $d$: $p_d = \frac{N_d}{\sum_{d'} N_{d'}},$
where $N_d$ is the number of prompts from domain $d$. 

We consider the following three scaling variants:
\begin{align}
\textbf{v1 (log):} \quad & w^{\text{dom}} = \log\left(1 + \frac{1}{p_d}\right) \\
\textbf{v2 (log-squared):} \quad & w^{\text{dom}} = \left[\log\left(1 + \frac{1}{p_d}\right)\right]^2 \\
\textbf{v3 (inverse):} \quad & w^{\text{dom}} = \frac{1}{p_d}
\end{align}


These variants are chosen to represent a spectrum of upweighting strengths. v1 (log) is hypothesized to provide a tempered yet significant boost to underrepresented domains; the logarithmic function naturally moderates the impact of extreme $p_d$ values, potentially enhancing training stability. v2 (log-squared) represents a more assertive non-linear scaling. v3 (inverse) offers the most direct and aggressive form of upweighting, making domain weights sharply inversely proportional to their frequency. This systematic variation from a more conservative (v1) to a highly aggressive (v3) approach allows us to study how different levels and types of domain correction affect training dynamics and model performance. A comparative analysis of these variants is presented in Section~\ref{sec:results:variant_selection}.

While domain-aware scaling helps rebalance across domains, it does not address intra-domain variation in prompt difficulty. Even within a single domain, some prompts are trivial while others are ambiguous. To account for this finer-grained challenge, we introduce difficulty-aware scaling.

\subsection{Difficulty-Aware Scaling}

While domain-aware weighting corrects for domain imbalance, it does not consider variation in prompt-level difficulty. As shown in empirical results (Section~\ref{sec:imbalance_results}), domains like IMDB achieve consistently high accuracy, whereas others such as Math yield lower accuracy, indicating that prompt complexity varies widely even under balanced training.

The GRPO algorithm provides a natural mechanism for estimating prompt-level difficulty. For each prompt \( q \), GRPO samples a group of \( G \) candidate completions \( \{o_i\}_{i=1}^G \), each associated with a binary reward \( r_i \in \{0, 1\} \). Previous work has shown that self-consistency, the agreement across model outputs, can serve as a proxy for uncertainty~\cite{wang2022self, li2023reflection}. Prompt groups with mixed outcomes reflect uncertainty and may benefit from greater optimization focus.

To capture this, we define the self-consistency (SC) score for prompt \( q \) as: $\text{SC}(q) = \frac{1}{G} \sum_{i=1}^{G} r_i,$
and define the difficulty weight as:
\begin{equation}
w^{\text{diff}}(q) = \frac{1}{\text{SC}(q) + \epsilon'},
\end{equation}
where \( \epsilon' \) is a small constant to ensure numerical stability.

Note that when all generations in a group are incorrect (i.e., \( \text{SC}(q) = 0 \)), $w^{\text{diff}}$ becomes large. However, this does not lead to instability, as all advantages will be zero based on Equation~\ref{eq:scale_adv}, and thus no policy update occurs. This mechanism encourages the model to focus more on prompts it finds uncertain, while ignoring uniformly poor or trivially easy cases.

Combining both components, we compute the final scaled reward of \ours as:
\begin{equation}
r_i^{\text{scaled}} = r_i \cdot w^{\text{dom}}(q) \cdot w^{\text{diff}}(q),
\end{equation}
where \( w^{\text{dom}}(q) \) and \( w^{\text{diff}}(q) \) are the domain- and difficulty-based weights for prompt group \( q \), respectively. The final advantage is then computed using Equation~\ref{eq:scale_adv}.
This formulation retains the structure of GRPO while enhancing it with principled scaling to reflect domain imbalance and prompt difficulty.

\section{Experiment Setup}
\label{sec:experiment}

We now evaluate the effectiveness of our \ours. Our experiments aim to assess whether these methods improve alignment performance on underrepresented or difficult domains, and whether the improvements are consistent across different models.

\paragraph{Dataset Setup.}
We evaluate across four task domains: \textsc{IMDB} (text classification), \textsc{GSM8K} and \textsc{MATH} (math problem solving), \textsc{Natural Questions} (open-domain QA), and \textsc{ARC} (reasoning QA). Our training dataset consists of 4,000 examples, as detailed in Table~\ref{tab:data_overview}.


\paragraph{Baseline Methods.}
We compare our method against the following baselines: (1) \textbf{Base Model}, the pretrained model without any fine-tuning; (2) \textbf{Naive GRPO}, which applies the original group-relative optimization without any reward reweighting; and (3) \textbf{Dr.~GRPO}~\cite{liu2025understanding}, which removes the length and standard deviation normalization. While other GRPO variants address issues like length bias or training instability~\cite{yu2503dapo, zhang2025srpo, zhang2025gvpo}, they are not included as baselines as our work targets GRPO's vulnerability to domain imbalance. In addition, we conduct ablation studies to isolate the contributions of domain-aware and difficulty-aware components.

\paragraph{Proposed Methods.}
We evaluate three variants of \ours, each utilizing one of the domain-aware weighting strategies in Section~\ref{sec:method_domain}. \textbf{\ours-Log} uses log-scaled domain weights (v1), 
\textbf{\ours-LogSq} uses squared-log weights (v2), and 
\textbf{\ours-Inv} uses inverse-frequency weights (v3). 
All variants incorporate the same difficulty-aware component based on self-consistency. This setup allows us to assess how different strengths of domain correction interact with difficulty-aware scaling in multi-domain alignment.

\paragraph{Model Setup.}
We evaluate our method across a diverse set of language models. These include \textbf{Qwen2.5-0.5B}, \textbf{Qwen2.5-1.5B}, \textbf{Qwen2-0.5B}, \textbf{Qwen3-0.6B}, \textbf{Qwen2.5-7B}, and \textbf{Qwen1.5-MoE-A2.7B} (14B total parameters, 2.7B activated) \cite{qwen2.5}, as well as \textbf{LLaMA3.2-1B} \cite{grattafiori2024llama}, \textbf{Olmo2-1B} \cite{olmo20242olmo2furious}, and \textbf{Gemma2-2B-it} \cite{gemma_2024}. This selection spans both dense and mixture-of-experts (MoE) architectures and covers a range of model capacities from 0.5B to 14B parameters.
For the GRPO group size, we use \( G = 2, 4, 8, 16 \) depending on the experiment. Additional implementation details are provided in Appendix~\ref{sec:implement}.

\section{Results}
\label{sec:result}

We begin our analysis by identifying the most effective variant of our proposed scaling strategy. After selecting a default method, we evaluate its impact through comparisons with baseline methods and targeted ablation studies.

\subsection{Identifying the Optimal Scaling Strategy} 
\label{sec:results:variant_selection} 

To determine the most effective variant of \ours, we compare three domain weighting strategies: \textbf{\ours-Log} (v1), \textbf{\ours-LogSq} (v2), and \textbf{\ours-Inv} (v3), across several models trained under domain-heavy conditions (group size $G = 4$). Table~\ref{tab:method_selection} summarizes these results. Each column shows the average accuracy across five evaluation datasets achieved when training under the specified domain-heavy condition. The final column (`Avg.') averages these scores across the four conditions.

\ours-Log achieves the highest average accuracy (`Avg.' column) across all tested models, striking the best balance between performance improvement and stability. While the more aggressive \ours-Inv or \ours-LogSq occasionally outperforms in specific settings (e.g., IMDB-heavy for LLaMA3.2-1B using v3), this often comes at the cost of reduced performance elsewhere, lowering their overall effectiveness. In contrast, \ours-Log delivers the top overall average score for all three models in this comparison. When averaged across all domain-model combinations shown in Table~\ref{tab:method_selection}, \ours-Log surpasses \ours-LogSq by $1.27\%$ and \ours-Inv by $0.97\%$. 

\ours-Log's superior performance, given that minority domains constitute only 8.3\% of our data, strongly aligns with logarithmic scaling's theoretical benefits. Its inherent diminishing returns temper responses to extreme domain characteristics, preventing over-correction and fostering a stable learning signal. In contrast, the aggressive scaling of \ours-LogSq and \ours-Inv, while occasionally beneficial in specific scenarios, led to higher performance variance and risked broader instability due to this pronounced data imbalance, thus explaining their lower average scores. Prioritizing robust alignment across domains through consistent and strong average performance, we therefore adopt \ours-Log for all subsequent experiments.

\begin{table}[btp]
\centering
\setlength{\tabcolsep}{2pt}
\renewcommand{\arraystretch}{1}
\resizebox{\linewidth}{!}{%
\begin{tabular}{lccccc}
\toprule
\textbf{Model} & \textbf{Math-heavy} & \textbf{IMDB-heavy} & \textbf{NQ-heavy} & \textbf{ARC-heavy} & \textbf{Avg.} \\
\midrule
\multicolumn{6}{c}{\textit{Qwen2.5-0.5B}} \\
\ours-Log       & \textbf{47.92} & 47.67 & \textbf{47.91} & \textbf{47.70} & \textbf{47.80} \\
\ours-LogSq     & 47.75 & 47.47 & 47.36 & 45.96 & 47.14 \\
\ours-Inv       & 47.72 & \textbf{47.81} & 47.59 & 46.22 & 47.33 \\
\midrule
\multicolumn{6}{c}{\textit{Qwen2.5-1.5B}} \\
\ours-Log       & 64.12 & 64.93 & \textbf{65.26} & \textbf{65.35} & \textbf{64.91} \\
\ours-LogSq     & \textbf{65.33} & \textbf{65.21} & 64.06 & 64.56 & 64.79 \\
\ours-Inv       & 65.26 & 64.47 & 64.67 & 64.92 & 64.83 \\
\midrule
\multicolumn{6}{c}{\textit{LLaMA3.2-1B}} \\
\ours-Log       & \textbf{28.97} & 30.02 & 22.19 & \textbf{30.56} & \textbf{27.94} \\
\ours-LogSq     & 27.49 & 25.51 & 22.46 & 24.22 & 24.92 \\
\ours-Inv       & 22.57 & \textbf{30.10} & \textbf{25.34} & 24.27 & 25.58 \\
\bottomrule
\end{tabular}
}
\vspace{-0.3cm}
\caption{\label{tab:method_selection} 
Average accuracy under each domain-heavy training setup for different reward scaling variants. Bold denotes the highest score for each condition.}
\vspace{-0.3cm}
\end{table}

\subsection{Comparison with Baselines}
\label{sec:results:baseline_comparison} 

\textbf{Our method outperforms baselines across different models and training distributions.} To evaluate the effectiveness of \ours, we compare \ours against two key baselines: Naive GRPO, the original formulation without reward rescaling, and Dr.~GRPO~\cite{liu2025understanding}. These comparisons utilize models trained ($G=4$) under the various domain-heavy distributions. We summarize the comparison in Table~\ref{tab:baseline_comparison}, which shows the average performance across all datasets under each setting. Complete results for individual datasets and additional base models, are shown in Appendix~\ref{sec:supp}, Table~\ref{tab:detailed_results_multirow} and Table~\ref{tab:additional_result}.

Focusing on Table~\ref{tab:baseline_comparison}, we first note that all GRPO variants evaluated consistently yield substantial improvements over their respective base models, confirming the effectiveness of GRPO-based alignment ~\cite{guo2025deepseek}. Turning to the comparison between the GRPO variants, we observe from the overall average performance (`Avg.' column) that \textbf{\ours achieves the highest overall average score on 5 out of the 6 models}. The most significant gains appear on the Qwen3-0.6B model, where \ours improves upon Naive GRPO by $4.40\%$ and upon Dr.~GRPO by $4.61\%$. Strong gains are also seen on LLaMA3.2-1B ($+1.65\%$ vs Naive, $+1.97\%$ vs Dr.~GRPO) and Qwen2.5-0.5B ($+1.63\%$ vs Naive, $+0.65\%$ vs Dr.~GRPO). Furthermore, \ours delivers the best average on the Qwen1.5-MoE ($+0.39\%$ vs Naive, $+1.44\%$ vs Dr.~GRPO). On Qwen2.5-1.5B, performance is very close, with Naive GRPO slightly ahead in the average ($64.97$ vs $64.91$ for \ours), though \ours outperforms both baselines under the NQ-heavy and ARC-heavy conditions individually. 

While Dr.~GRPO occasionally surpasses Naive GRPO, the consistent advantages of \ours, most notably the substantial gains on Qwen3-0.6B, LLaMA3.2-1B, and Qwen2.5-0.5B, underscore the effectiveness of incorporating explicit domain and difficulty signals. Statistical testing further confirms this improvement: compared to Naive GRPO, \ours achieved a $t$-statistic of $2.97$ with a one-tailed $p$-value of $0.0018$, and compared to Dr.~GRPO, it yielded a $t$-statistic of $3.49$ with a one-tailed $p$-value of $0.0003$ \cite{gould2010statistics}. Together, these results highlight the robustness of our joint scaling approach in enhancing GRPO alignment, particularly in mitigating performance trade-offs introduced by domain imbalance.

\begin{table}[tbp]
\centering
\setlength{\tabcolsep}{2pt}
\renewcommand{\arraystretch}{1}
\resizebox{\linewidth}{!}{%
\begin{tabular}{lccccc}
\toprule
\textbf{Model} & \textbf{Math-heavy} & \textbf{IMDB-heavy} & \textbf{NQ-heavy} & \textbf{ARC-heavy} & \textbf{Avg.} \\
\midrule
\multicolumn{6}{c}{\textit{Qwen2.5-0.5B}} \\
Base           & 42.80          &               &               &               &               \\
Naive GRPO     & 46.22          & 46.21          & 46.43          & 45.80          & 46.17          \\
Dr GRPO        & 46.90          & 46.22          & 47.84          & 47.62          & 47.14          \\ 
\ours       & \textbf{47.92} & \textbf{47.67} & \textbf{47.91} & \textbf{47.70} & \textbf{47.80} \\
\midrule
\multicolumn{6}{c}{\textit{Qwen2.5-1.5B}} \\
Base           & 60.06          &               &               &               &               \\
Naive GRPO     & \textbf{64.81} & 64.91          & 65.15          & 65.02          & \textbf{64.97} \\
Dr GRPO        & 64.76          & 64.75          & 64.72          & 65.16          & 64.85          \\
\ours       & 64.12          & \textbf{64.93} & \textbf{65.26} & \textbf{65.35} & 64.91          \\ 
\midrule
\multicolumn{6}{c}{\textit{Qwen2-0.5B}} \\
Base           & 43.18          &               &               &               &               \\
Naive GRPO     & 44.05          & 43.76          & 43.97          & \textbf{45.35} & 44.28          \\
Dr GRPO        & \textbf{45.48} & 44.31          & 43.14          & 44.87          & 44.45          \\ 
\ours       & 43.98          & \textbf{44.83} & \textbf{44.50} & 45.07          & \textbf{44.60} \\
\midrule
\multicolumn{6}{c}{\textit{LLaMA3.2-1B}} \\
Base           & 22.31          &               &               &               &               \\
Naive GRPO     & 27.61          & 29.94          & 21.76          & 25.86          & 26.29          \\ 
Dr GRPO        & 22.24          & 29.72          & 22.17          & 29.74          & 25.97          \\
\ours       & \textbf{28.97} & \textbf{30.02} & \textbf{22.19} & \textbf{30.56} & \textbf{27.94} \\
\midrule
\multicolumn{6}{c}{\textit{Qwen1.5-MoE}} \\
Base           & 45.57          &               &               &               &               \\
Naive GRPO     & 66.05          & 64.03          & \textbf{66.04} & 65.46          & 65.40          \\
Dr GRPO        & 62.46          & 64.85          & 64.50          & 65.58          & 64.35          \\ 
\ours       & \textbf{66.37} & \textbf{65.59} & 65.19          & \textbf{66.00} & \textbf{65.79} \\
\midrule
\multicolumn{6}{c}{\textit{Qwen3-0.6B}} \\
Base           & 25.59          &               &               &               &               \\
Naive GRPO     & 41.99          & 34.83          & 34.72 & 34.31          &   36.46        \\
Dr GRPO        & 42.08          & 33.23          & 36.76          & 32.91          &    36.25      \\ 
\ours       & \textbf{45.07} & \textbf{35.64} & \textbf{44.44}          & \textbf{38.28} & \textbf{40.86} \\
\bottomrule
\end{tabular}
}
\vspace{-0.3cm}
\caption{\label{tab:baseline_comparison}
Average accuracy under each domain-heavy training setup for different methods. Scores are averaged over five task-specific datasets per domain. Bold indicates the best-performing method.} 
\end{table}

\begin{figure}[t] 
\centering 
\includegraphics[width=\linewidth]{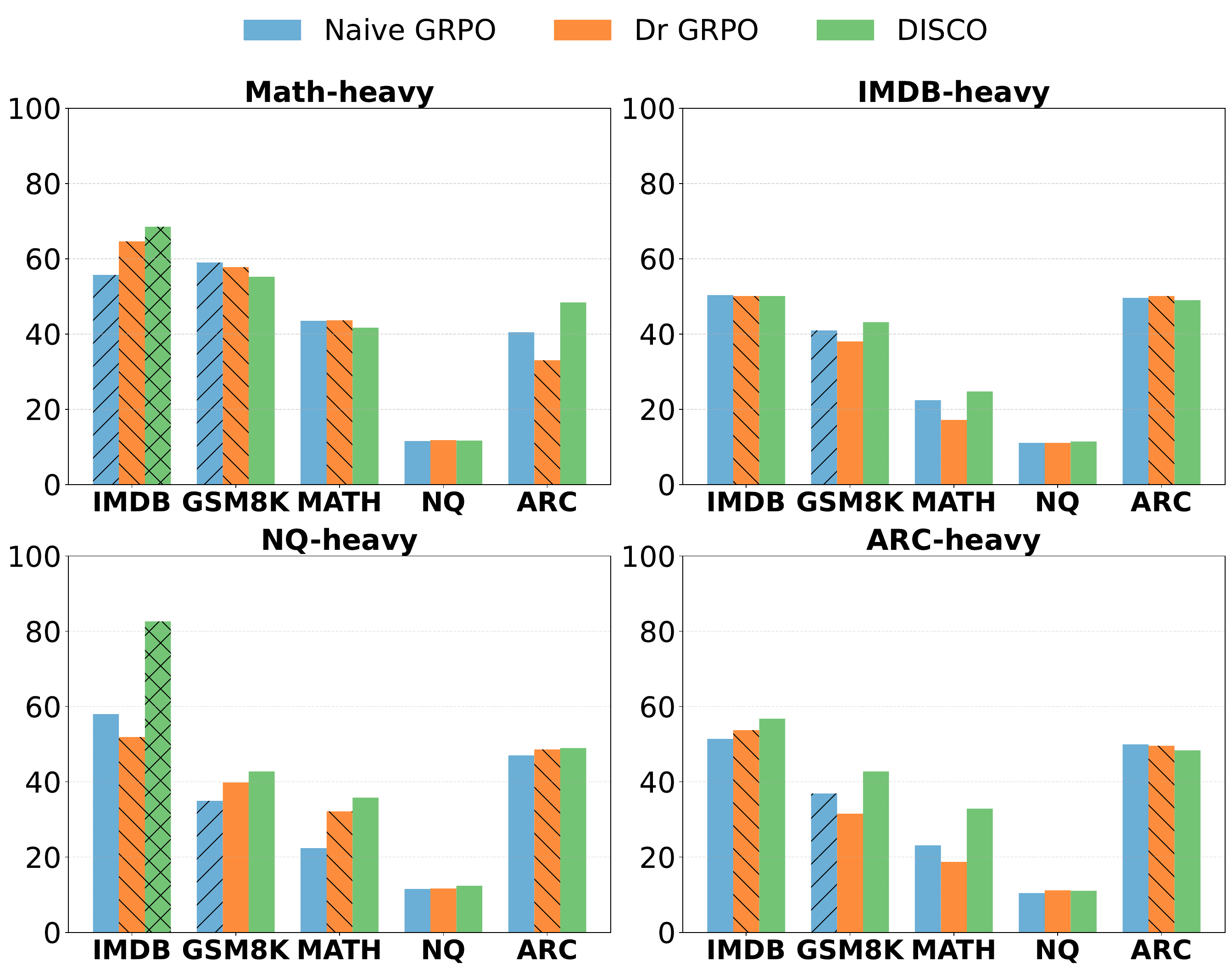} 
\caption{Qwen3-0.6B performance (EM Accuracy \%) breakdown by the dataset under four domain-heavy training conditions (one condition per panel/subplot). Bar groups show results on individual datasets.}
\label{fig:qwen3_breakdown_barplots} 
\end{figure}

\paragraph{Performance Breakdown by Dataset.}
\label{sec:results:breakdown} 
To illustrate how our method navigates domain trade-offs, we break down Qwen3-0.6B performance by evaluation dataset under each domain-heavy setting (Figure~\ref{fig:qwen3_breakdown_barplots}). This case study compares Naive GRPO, Dr.~GRPO, and \ours on the individual datasets.

The results for Qwen3-0.6B reveal a clear pattern where \ours significantly boosts minority domain performance, sometimes involving a trade-off with majority domain scores. On majority domains (those comprising 75\% of training data), \ours performance varies compared to Naive GRPO. For instance, when trained NQ-heavy, \ours improves performance on the NQ dataset ($12.30\%$ vs $11.47\%$). However, when trained Math-heavy, it scores lower on both MATH ($41.68\%$ vs $43.40\%$) and GSM8K ($55.20\%$ vs $58.91\%$). Similarly, under ARC-heavy training, the ARC score is slightly lower ($48.29\%$ vs $49.91\%$).

In contrast, \ours demonstrates substantial and consistent improvements on tail domains. The most dramatic gains are seen in the NQ-heavy setting: the IMDB score jumps to $82.60\%$ (from $57.90\%$ for Naive GRPO), GSM8K increases to $42.72\%$ (from $34.86\%$), and MATH rises to $35.70\%$ (from $22.34\%$). Significant minority  domain recovery is also evident in other settings, such as the gains on GSM8K ($+2.28\%$) and MATH ($+2.34\%$) under IMDB-heavy training.

These Qwen3-0.6B results show our joint scaling strategy effectively counters dominant domain overfitting, leading to significant recovery on minority domains, sometimes at the cost of peak head-domain performance. This yields a more balanced, generalized performance profile across diverse tasks, driven by the synergy between domain-aware reweighting and difficulty-aware scaling.

\subsection{Ablation Study of Scaling Components}
\label{sec:results:ablation} 

\textbf{Combining domain- and difficulty-aware scaling delivers the strongest overall performance.} To understand the individual contributions of our proposed scaling strategies, we conduct an ablation study selectively applying either the domain-aware weight (`Domain only') or the difficulty-aware weight (`Diff only'), comparing against the full version (`\ours') and the Naive GRPO baseline. Results are reported in Table~\ref{tab:ablation_study}.

\begin{table}[tbp]
\centering
\setlength{\tabcolsep}{2pt}
\renewcommand{\arraystretch}{1}
\resizebox{\linewidth}{!}{%
\begin{tabular}{@{}lccccc@{}} 
\toprule
\textbf{Model} & \textbf{Math-heavy} & \textbf{IMDB-heavy} & \textbf{NQ-heavy} & \textbf{ARC-heavy} & \textbf{Avg.} \\ 
\midrule
\multicolumn{6}{c}{\textit{Qwen2.5 0.5B}} \\
 Naive GRPO         & 46.22          & 46.21          & 46.43          & 45.80          & 46.17          \\ 
 Diff only          & 46.87          & 46.21          & 45.03          & 46.07          & 46.05          \\ 
 Domain only        & 45.94          & 46.33          & 45.28          & 46.10          & 45.91          \\ 
 \textbf{\ours}  & \textbf{47.92} & \textbf{47.67} & \textbf{47.91} & \textbf{47.70} & \textbf{47.80} \\ 
\midrule
\multicolumn{6}{c}{\textit{Qwen2.5 1.5B}} \\
 Naive GRPO         & 64.81          & 64.91          & 65.15          & 65.02          & 64.97          \\ 
 Diff only          & 64.45          & 64.40          & 64.67          & \textbf{65.39} & 64.73          \\ 
 Domain only        & \textbf{65.39} & 64.73          & 64.65          & 65.18          & \textbf{64.99} \\ 
 \textbf{\ours}  & 64.12          & \textbf{64.93} & \textbf{65.26} & 65.35          & 64.92          \\ 
\midrule
\multicolumn{6}{c}{\textit{LLaMA3.2 1B}} \\
 Naive GRPO         & 27.61          & 29.94          & 21.76          & 25.86          & 26.29          \\ 
 Diff only          & 28.67          & 29.87          & \textbf{22.46} & 30.52          & 27.88          \\ 
 Domain only        & 26.42          & 29.61          & 21.87          & 30.01          & 26.98          \\ 
 \textbf{\ours}  & \textbf{28.97} & \textbf{30.02} & 22.19          & \textbf{30.56} & \textbf{27.94} \\ 
\bottomrule
\end{tabular}
}
\caption{\label{tab:ablation_study}
Ablation study comparing Naive GRPO with variants using only difficulty-aware scaling (`Diff only'), only domain-aware log-scaling (`Domain only', v1 weights), and both (`\ours'). Best result among the three variants in each numerical column is bolded.}
\end{table}

We observe varied effects from the individual components across models. For instance, on Qwen2.5-0.5B, neither difficulty-aware scaling alone ($46.05\%$) nor domain-aware scaling alone ($45.91\%$) improved upon Naive GRPO ($46.17\%$) on average. However, combining both in \ours yields a significant boost to $47.80\%$. Conversely, on LLaMA3.2-1B, both `Diff only' ($27.88\%$) and `Domain only' ($26.98\%$) offer improvements over Naive GRPO ($26.29\%$), and  \ours achieves the highest overall score ($27.94\%$).

Interestingly, on the larger Qwen2.5-1.5B model, using `Domain only' scaling achieves the highest average score ($64.99\%$), slightly surpassing both Naive GRPO ($64.97\%$) and \ours ($64.92\%$). This suggests domain re-weighting alone can be particularly effective for this model configuration. Despite `Domain only' having the best average here, \ours ($64.92\%$) achieves competitive overall performance and secures the best scores among the variants under the IMDB-heavy and NQ-heavy conditions specifically.

These results reveal a complementary relationship between the two scaling components. Relying on one component alone proves insufficient: `Domain only' scaling neglects sample difficulty variance within domains, while `Diff only' scaling ignores global domain imbalance, both leading to inconsistent performance. In contrast, their joint use in \ours consistently provides a more robust and generally higher-performing configuration across different models and domain imbalances, validating our combined approach.

\subsection{Effect of Group Size on Alignment Quality}

\begin{figure}[t]
\centering
\includegraphics[width=\linewidth]{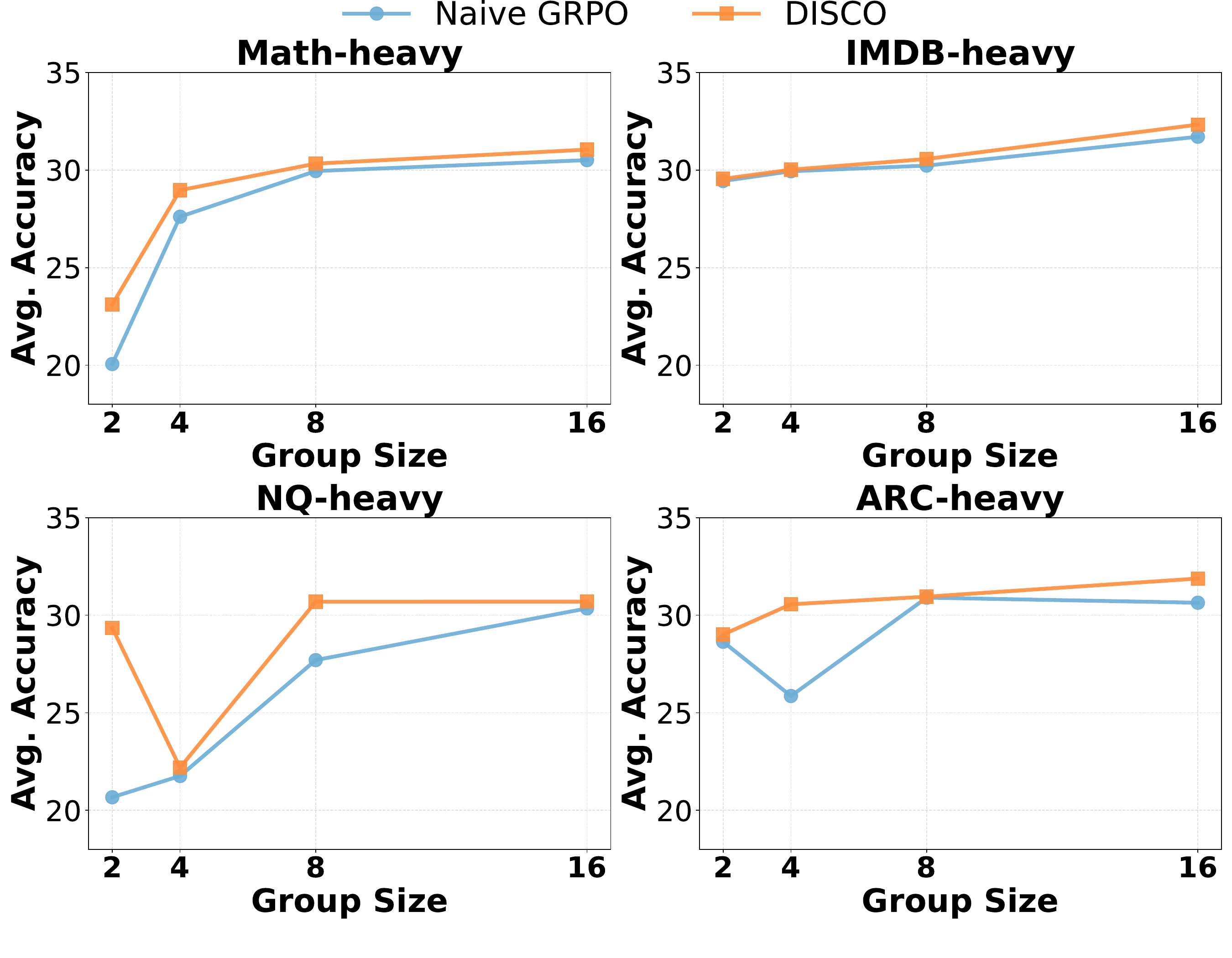}
\vspace{-0.8cm}
\caption{Impact of group size on LLaMA3.2-1B performance (EM Accuracy \%) under different domain-heavy training distributions. Each line shows results for Naive GRPO and \ours as group size varies.}
\label{fig:group_size_lineplots}
\end{figure}

To further test the robustness of \ours, we vary the group size $G$ used during GRPO and evaluate its impact. Figure~\ref{fig:group_size_lineplots} shows the performance of Naive GRPO and \ours across $G=\{2,4,8,16\}$ on LLaMA3.2-1B under different training settings.

\textbf{Our method generalizes well across different group sizes, outperforming the baseline.} Across all setups, we observe that increasing group size generally improves performance for both methods. \ours consistently outperforms Naive GRPO across almost all group sizes, highlighting its strong generalization ability. For example, under the Math-heavy setup, \ours improves the average score from $20.06\%$ to $23.11\%$ at group size 2, and from $30.51\%$ to $31.05\%$ at group size 16. Similarly, in the ARC-heavy setup, \ours reaches $31.88\%$ compared to $30.64\%$ for Naive GRPO at the largest group size. These results demonstrate that the benefits of reward scaling are robust to the group size $G$, making the approach broadly applicable within GRPO frameworks. Furthermore, this robustness extends to dataset size variations, as our method maintains consistent advantages when trained on 2,000 examples, detailed in Appendix~\ref{sec:appendix_datasize}.

\section{Related Work}

\noindent \textbf{Reinforcement Learning from Human Feedback (RLHF)} has emerged as a powerful paradigm for aligning LLMs with human preferences and values. The most popular approach is Proximal Policy Optimization (PPO)~\cite{ouyang2022training, schulman2017proximal}. Building upon PPO, GRPO eliminates the need for value functions through estimation from group scores and has emerged as a popular training diagram~\cite{shao2024deepseekmath}. Rule-based GRPO further streamlines training by using predefined rules to evaluate outputs, avoiding the need for a separate reward model~\cite{guo2025deepseek}. Variants address specific limitations of GRPO have also been proposed \cite{su2025crossing}. To mitigate length bias, Dr. GRPO removes normalization terms while SRPO introduces two-stage training and historical sampling~\cite{liu2025understanding, zhang2025srpo}. To address training instability,  DAPO increases clip rates, and GVPO offers an analytical solution to KL-constrained reward maximization~\cite{yu2503dapo, zhang2025gvpo}.  While prior work focuses on training challenges, we examine its vulnerability to dataset imbalance. Our proposed strategies are complementary and can be integrated with existing GRPO variants for improved performance.



\noindent \textbf{Data Imbalance in NLP} is a pervasive challenge characterized by significantly skewed distributions in training data. This issue is prevalent across various NLP tasks and often results in models that are biased toward the majority class, leading to poor performance on minority classes~\cite{henning2023survey, he2009learning, liu2019large, wang2017learning, godbole2022benchmarking, dai2023long, liu2024large}. Approaches to addressing data imbalance include data augmentation and loss adjustment methods~\cite{zhou2024multi, sun2007cost, kandpal2023large, zhang2022making, tepper2020balancing, li2020dice, kuo2024not}. Our work aligns with the loss adjustment paradigm. However, instead of reweighting loss, we adjust the rewards used in GRPO. While data augmentation techniques, such as generating synthetic data, can rebalance data distributions, they often introduce additional costs. In contrast, we focus on developing policy modifications to the GRPO process without incurring extra costs. Notably, \ours can be combined with data augmentation techniques.

\section{Conclusion}
In this work, we investigate the vulnerability of GRPO to domain imbalance in multi-domain alignment and introduce \ours that combines domain and difficulty-aware reward scaling to enable GRPO to better handle skewed training distributions. Through experiments across diverse model scales, group sizes, and alignment settings, we demonstrate that \ours consistently improves generalization and outperforms existing baselines. Ablation studies further confirm that both scaling components are essential. Overall, our method provides a robust enhancement to GRPO for aligning LLMs with real-world, imbalanced data.

\section{Limitation}
While our proposed domain- and difficulty-aware scaling demonstrates promising results for enhancing GRPO's robustness to domain imbalance, this study has several limitations that suggest avenues for future research. Firstly, our experiments primarily utilized deterministic, rule-based rewards (Exact Match) to isolate the effects of the optimization dynamics. Real-world RLHF often involves complex, potentially noisy, and continuous rewards learned from human preferences. The efficacy and stability of our scaling methods under such learned reward functions warrant further investigation.

Secondly, we assumed the availability of clear, predefined domain labels for applying domain-aware scaling. The practical application to datasets with ambiguous, overlapping, or fine-grained domain structures, potentially requiring automatic domain identification, remains an open challenge. Furthermore, while we evaluated across diverse architectures and sizes up to the 14B scale (MoE), verification of performance consistency on significantly larger foundation models may still be needed.

Our methodological choices also present areas for future exploration. The reliance on self-consistency within a sampled group as a proxy for prompt difficulty is one of several possible heuristics. This work focused exclusively on modifying the GRPO algorithm's optimization dynamics through reward scaling and did not explore alternative or potentially complementary strategies for handling domain imbalance that operate at the data level, such as data augmentation techniques (e.g., oversampling rare domains or generating synthetic data). Methods like these could offer different trade-offs and could potentially be combined with our algorithmic approach. Moreover, the interaction between our scaling mechanisms and other crucial RLHF hyperparameters, such as the KL divergence coefficient or learning rates, was not exhaustively studied.

Finally, our current evaluation was based on four predefined domain-heavy training settings, and performance was typically assessed using an unweighted average across different domain datasets. Future research could explore a more continuous spectrum of domain imbalance ratios by dynamically sampling data to construct a wider variety of training datasets. This would also allow for evaluation using different weighted averaging schemes for domain-specific scores, enabling a more nuanced understanding and potentially the mapping of Pareto-optimal frontiers in multi-domain performance.

\section{Acknowledgment}
Zhou, Liu, Wang and Huang are supported by DARPA Transfer from Imprecise and Abstract Models to Autonomous Technologies (TIAMAT) 80321, DARPA HR001124S0029-AIQ-FP-019, DOD-AFOSR-Air Force Office of Scientific Research under award number FA9550-23-1-0048, National Science Foundation NSF-IIS-2147276FAI, National Science Foundation NAIRR240045, National Science Foundation TRAILS Institute (2229885). Private support was provided by Peraton.

\bibliographystyle{acl_natbib}
\bibliography{custom}

\appendix
\section{Implementation Details}
\label{sec:implement}

We train models using the \texttt{OpenRLHF} framework \cite{hu2024openrlhf} with GRPO optimization. Each model is fine-tuned for 1 epoch. The rollout and training batch sizes are both set to 64, with micro-batch sizes of 8 and 4, respectively. We use a maximum prompt and generation length of 1024 tokens. The KL penalty is initialized at $1\mathrm{e}{-3}$ and estimated using the K3 estimator. All models use a learning rate of $1\mathrm{e}{-6}$.

All evaluations are conducted using zero-shot inference, with models generating answers without access to in-context examples. For each task, we apply a fixed, task-specific prompt template for both training and evaluation. The prompts are designed to clearly define the task instruction and input format.

Table~\ref{tab:prompt_templates} presents the prompt templates used across the five datasets.

\begin{table}[htbp]
\centering
\resizebox{\linewidth}{!}{%
\begin{tabular}{l|p{14cm}} 
\toprule
\textbf{Dataset} & \textbf{Prompt Template} \\
\midrule
\textbf{MATH} & 
\texttt{Below is a math problem. Provide a detailed, step-by-step solution.} \texttt{\#\#\# Problem: \{problem\}} \texttt{\#\#\# Answer:} \\ 
\midrule
\textbf{GSM8K} & 
Same as MATH \\
\midrule
\textbf{IMDB} & 
\texttt{Below is a movie review. Determine the sentiment of the review as Positive or Negative.} \texttt{\#\#\# Review: \{review\}} \texttt{\#\#\# Answer:} \\ 
\midrule
\textbf{NQ} & 
\texttt{Below is a question that requires a concise and accurate answer. Provide a detailed explanation before concluding with the correct answer.} \texttt{\#\#\# Question: \{question\}} \texttt{\#\#\# Answer:} \\ 
\midrule
\textbf{ARC} & 
\texttt{Below is a question with multiple-choice answers. Choose the correct option based on your reasoning.} \texttt{\#\#\# Question: \{question\}} \texttt{\#\#\# Choices:} \texttt{A. \{choice\_A\}} \texttt{B. \{choice\_B\}} \texttt{...} \texttt{\#\#\# Answer:} \\ 
\bottomrule
\end{tabular}
}
\caption{\label{tab:prompt_templates} Prompt templates used for zero-shot inference. Each template is applied consistently during both training and evaluation.}
\end{table}

\section{Effect of Reduced Dataset Size}
\label{sec:appendix_datasize}

To assess the robustness of our proposed scaling method to varying amounts of alignment data, we conducted additional experiments using a smaller dataset size of 2,000 examples (compared to 4,000 in the main experiments). We evaluated performance on the Qwen3-0.6B model under the four domain-heavy training distributions (G=4).

The detailed results are presented in Table~\ref{tab:detailed_results_qwen3_2k}. Comparing the average performance (`Avg.' column) across the heavy training settings, we find that \ours consistently outperforms both Naive GRPO and Dr. GRPO even with this reduced dataset. Averaged across the four heavy conditions, our method achieves a score of $33.97\%$, representing gains of $+3.47\%$ over Naive GRPO ($30.50\%$) and $+3.20\%$ over Dr. GRPO ($30.77\%$).

Notably, significant improvements are still observed on tail domains. For instance, under IMDB-heavy training, \ours substantially improves GSM8K ($+12.05\%$ vs Naive) and MATH ($+3.40\%$ vs Naive) performance. While the absolute scores are generally lower than those achieved with 4,000 data (as expected), the relative advantage and the effectiveness of the domain- and difficulty-aware scaling mechanism persist. This supports the generalization ability of our method across different dataset sizes.

\begin{table}[htbp] 
\centering
\resizebox{\linewidth}{!}{%
\begin{tabular}{@{}ll rrrrrr@{}} 
\toprule
\textbf{Training Dist.} & \textbf{Method} & \textbf{IMDB} & \textbf{GSM8K} & \textbf{MATH} & \textbf{NQ} & \textbf{ARC} & \textbf{Avg.} \\
\midrule
\multicolumn{8}{c}{\textit{Qwen3-0.6B (2K Training Data)}} \\ 
\multirow{3}{*}{Math heavy} & Naive GRPO         & 62.00          & \textbf{52.72} & 39.66          & 11.05          & 37.12          & 40.51          \\ 
                             & Dr. GRPO           & 67.80          & 50.98          & \textbf{41.18} & 10.99          & 35.58          & 41.31          \\ 
                             & \ours  & \textbf{75.60} & 49.67          & 33.96          & \textbf{11.33} & \textbf{47.01} & \textbf{43.51} \\ 
\midrule
\multirow{3}{*}{IMDB heavy} & Naive GRPO         & 52.20          & 12.07          & 11.32          & 10.50          & 43.94          & 26.01          \\ 
                             & Dr. GRPO           & 51.20          & 11.21          & 12.04          & 10.72          & 44.28          & 25.89          \\ 
                             & \ours  & \textbf{55.50} & \textbf{24.12} & \textbf{14.72} & \textbf{11.05} & \textbf{46.58} & \textbf{30.39} \\ 
\midrule
\multirow{3}{*}{NQ heavy}   & Naive GRPO         & 57.90          & 11.03          & 10.50          & \textbf{11.91} & 44.11          & 26.79          \\ 
                             & Dr. GRPO           & 56.40          & 12.82          & 11.00          & 11.55          & 44.62          & 27.28          \\ 
                             & \ours  & \textbf{82.60} & \textbf{20.02} & \textbf{13.40} & 10.94          & \textbf{46.58} & \textbf{34.71} \\ 
\midrule
\multirow{3}{*}{ARC heavy}  & Naive GRPO         & 54.40          & 13.96          & 13.00          & \textbf{11.14} & \textbf{51.02} & 28.70          \\ 
                             & Dr. GRPO           & 54.40          & 14.44 & 12.60          & 11.11          & 50.34          & 28.58          \\ 
                             & \ours  & \textbf{55.20} & \textbf{32.66} & \textbf{19.42} & 11.05          & 48.37          & \textbf{33.34} \\ 
\bottomrule
\end{tabular}
}
\caption{Detailed results for Qwen3-0.6B trained with a reduced dataset size (2K examples, G=4). Scores are EM accuracy (\%) on individual datasets and the average, comparing alignment methods across domain-heavy training distributions.}
\label{tab:detailed_results_qwen3_2k} 
\end{table}

\section{Supplementary Material}
\label{sec:supp}

\subsection{Detailed Results on Individual Datasets}
Detailed results for Table \ref{tab:baseline_comparison} in Section \ref{sec:results:baseline_comparison} are shown in Table \ref{tab:detailed_results_multirow}.

\begin{table}[htbp] 
\centering
\resizebox{\linewidth}{!}{%
\begin{tabular}{@{}ll rrrrrr@{}} 
\toprule
\textbf{Training Dist.} & \textbf{Method} & \textbf{IMDB} & \textbf{GSM8K} & \textbf{MATH} & \textbf{NQ} & \textbf{ARC} & \textbf{Avg.} \\
\midrule
\multicolumn{8}{c}{\textit{Qwen2.5 0.5B}} \\ 
\multirow{3}{*}{Math heavy} & Naive GRPO         & 87.80 & 58.98 & 24.38 & 13.96 & 45.99 & 46.22 \\
                             & \ours-Log       & 90.40 & 59.32 & 29.68 & 13.85 & 46.33 & 47.92 \\
                             & Dr. GRPO           & 90.30 & 60.24 & 24.56 & 13.91 & 45.50 & 46.90 \\
\multirow{3}{*}{IMDB heavy} & Naive GRPO         & 90.90 & 56.88 & 23.54 & 13.85 & 45.90 & 46.21 \\
                             & \ours-Log       & 89.60 & 57.88 & 29.86 & 14.43 & 46.59 & 47.67 \\
                             & Dr. GRPO           & 91.30 & 56.96 & 23.54 & 13.99 & 45.30 & 46.22 \\
\multirow{3}{*}{NQ heavy}   & Naive GRPO         & 89.40 & 56.22 & 23.36 & 16.57 & 46.58 & 46.43 \\
                             & \ours-Log       & 91.30 & 57.29 & 29.92 & 15.29 & 45.73 & 47.91 \\
                             & Dr. GRPO           & 90.40 & 56.65 & 29.26 & 16.89 & 45.98 & 47.84 \\ 
\multirow{3}{*}{ARC heavy}  & Naive GRPO         & 88.80 & 55.29 & 22.46 & 14.16 & 48.29 & 45.80 \\
                             & \ours-Log       & 90.60 & 56.48 & 29.44 & 14.04 & 47.95 & 47.70 \\ 
                             & Dr. GRPO           & 90.60 & 56.12 & 29.54 & 13.32 & 48.54 & 47.62 \\ 
\midrule

\multicolumn{8}{c}{\textit{Qwen2.5 1.5B}} \\
\multirow{3}{*}{Math heavy} & Naive GRPO         & 93.80 & 84.60 & 52.34 & 25.98 & 67.32 & 64.81 \\
                             & \ours-Log       & 93.10 & 84.02 & 51.90 & 26.57 & 65.02 & 64.12 \\
                             & Dr. GRPO           & 92.90 & 84.76 & 52.16 & 25.40 & 68.56 & 64.76 \\
\multirow{3}{*}{IMDB heavy} & Naive GRPO         & 94.10 & 83.58 & 51.30 & 25.87 & 69.71 & 64.91 \\
                             & \ours-Log       & 93.20 & 83.54 & 50.86 & 25.10 & 71.93 & 64.93 \\
                             & Dr. GRPO           & 93.20 & 83.78 & 51.10 & 24.56 & 71.10 & 64.75 \\
\multirow{3}{*}{NQ heavy}   & Naive GRPO         & 93.60 & 83.52 & 50.80 & 28.28 & 69.54 & 65.15 \\
                             & \ours-Log       & 92.90 & 83.50 & 50.32 & 27.76 & 71.84 & 65.26 \\
                             & Dr. GRPO           & 92.60 & 83.41 & 50.26 & 27.56 & 69.79 & 64.72 \\
\multirow{3}{*}{ARC heavy}  & Naive GRPO         & 93.40 & 83.44 & 50.70 & 24.45 & 73.12 & 65.02 \\
                             & \ours-Log       & 93.10 & 83.22 & 50.40 & 26.57 & 73.46 & 65.35 \\
                             & Dr. GRPO           & 93.30 & 83.45 & 50.49 & 25.42 & 73.12 & 65.16 \\
\midrule

\multicolumn{8}{c}{\textit{Qwen2-0.5B}} \\ 
\multirow{3}{*}{Math heavy} & Naive GRPO         & 90.80 & 76.53 &  6.28 & 12.66 & 33.96 & 44.05 \\
                             & \ours-Log       & 92.00 & 76.23 &  6.38 & 13.38 & 31.91 & 43.98 \\
                             & Dr. GRPO           & 91.70 & 76.90 &  6.16 & 13.21 & 39.42 & 45.48 \\
\multirow{3}{*}{IMDB heavy} & Naive GRPO         & 91.70 & 72.90 &  7.44 & 13.68 & 33.11 & 43.76 \\
                             & \ours-Log       & 90.40 & 74.66 &  7.64 & 13.46 & 37.97 & 44.83 \\
                             & Dr. GRPO           & 90.50 & 74.29 &  7.00 & 12.69 & 37.05 & 44.31 \\
\multirow{3}{*}{NQ heavy}   & Naive GRPO         & 91.70 & 72.51 &  7.92 & 14.29 & 33.45 & 43.97 \\
                             & \ours-Log       & 91.50 & 74.70 &  7.20 & 13.49 & 35.60 & 44.50 \\
                             & Dr. GRPO           & 91.50 & 72.54 &  8.02 & 13.85 & 29.78 & 43.14 \\
\multirow{3}{*}{ARC heavy}  & Naive GRPO         & 91.20 & 74.12 &  6.90 & 12.99 & 41.55 & 45.35 \\
                             & \ours-Log       & 90.70 & 74.27 &  6.68 & 13.16 & 40.53 & 45.07 \\
                             & Dr. GRPO           & 89.70 & 72.63 &  8.18 & 12.35 & 41.51 & 44.87 \\
\midrule

\multicolumn{8}{c}{\textit{Qwen1.5-MoE-A2.7B}} \\ 
\multirow{3}{*}{Math heavy} & Naive GRPO         & 92.80 & 91.22 & 29.16 & 44.88 & 72.18 & 66.05 \\
                             & \ours-Log       & 93.70 & 92.05 & 29.46 & 43.68 & 72.95 & 66.37 \\
                             & Dr. GRPO           & 94.50 & 92.06 & 29.42 & 45.65 & 50.68 & 62.46 \\
\multirow{3}{*}{IMDB heavy} & Naive GRPO         & 95.40 & 83.50 & 24.80 & 43.57 & 72.86 & 64.03 \\
                             & \ours-Log       & 94.90 & 87.22 & 26.30 & 46.04 & 73.46 & 65.59 \\
                             & Dr. GRPO           & 94.10 & 87.16 & 25.74 & 45.15 & 72.10 & 64.85 \\
\multirow{3}{*}{NQ heavy}   & Naive GRPO         & 94.50 & 90.03 & 26.84 & 46.48 & 72.35 & 66.04 \\
                             & \ours-Log       & 94.80 & 90.24 & 27.26 & 44.70 & 68.94 & 65.19 \\
                             & Dr. GRPO           & 94.40 & 85.41 & 24.28 & 46.39 & 72.01 & 64.50 \\
\multirow{3}{*}{ARC heavy}  & Naive GRPO         & 94.70 & 86.61 & 27.20 & 43.88 & 74.91 & 65.46 \\
                             & \ours-Log       & 94.70 & 90.34 & 27.16 & 43.93 & 73.91 & 66.00 \\
                             & Dr. GRPO           & 94.70 & 88.44 & 26.22 & 45.31 & 73.72 & 65.68 \\
\midrule

\multicolumn{8}{c}{\textit{Llama3.2 1B}} \\ 
\multirow{3}{*}{Math heavy} & Naive GRPO         & 75.60 &  3.25 &  3.94 & 21.72 & 33.53 & 27.61 \\
                             & \ours-Log       & 85.10 &  3.06 &  3.06 & 21.14 & 32.51 & 28.97 \\
                             & Dr. GRPO           & 50.70 &  2.79 &  3.16 & 22.02 & 32.51 & 22.24 \\
\multirow{3}{*}{IMDB heavy} & Naive GRPO         & 87.30 &  3.86 &  4.02 & 19.97 & 34.56 & 29.94 \\
                             & \ours-Log       & 88.80 &  3.14 &  3.34 & 20.97 & 33.87 & 30.02 \\
                             & Dr. GRPO           & 88.00 &  3.54 &  3.50 & 20.47 & 33.07 & 29.72 \\
\multirow{3}{*}{NQ heavy}   & Naive GRPO         & 50.30 &  2.85 &  2.96 & 23.19 & 29.52 & 21.76 \\
                             & \ours-Log       & 51.10 &  2.54 &  3.08 & 22.66 & 31.57 & 22.19 \\
                             & Dr. GRPO           & 50.90 &  2.87 &  2.88 & 22.29 & 31.93 & 22.17 \\
\multirow{3}{*}{ARC heavy}  & Naive GRPO         & 66.60 &  2.69 &  2.98 & 20.19 & 36.86 & 25.86 \\
                             & \ours-Log       & 87.90 &  3.31 &  3.74 & 19.02 & 38.83 & 30.56 \\
                             & Dr. GRPO           & 86.10 &  3.40 &  3.34 & 19.78 & 36.09 & 29.74 \\
\midrule
\multicolumn{8}{c}{\textit{Qwen3-0.6B}} \\ 
\multirow{3}{*}{Math heavy} & Naive GRPO         & 55.70 & 58.91 & 43.40 & 11.51 & 40.44 & 41.99 \\ 
                             & \ours-Log       & 68.50 & 55.20 & 41.68 & 11.61 & 48.38 & 45.07 \\ 
                             & Dr. GRPO           & 64.50 & 57.69 & 43.54 & 11.72 & 32.94 & 42.08 \\ 
\multirow{3}{*}{IMDB heavy} & Naive GRPO         & 50.30 & 40.85 & 22.40 & 11.05 & 49.57 & 34.83 \\ 
                             & \ours-Log       & 50.10 & 43.13 & 24.74 & 11.36 & 48.89 & 35.64 \\ 
                             & Dr. GRPO           & 50.00 & 37.97 & 17.06 & 11.05 & 50.09 & 33.23 \\ 
\multirow{3}{*}{NQ heavy}   & Naive GRPO         & 57.90 & 34.86 & 22.34 & 11.47 & 47.01 & 34.72 \\ 
                             & \ours-Log       & 82.60 & 42.72 & 35.70 & 12.30 & 48.89 & 44.44 \\ 
                             & Dr. GRPO           & 51.90 & 39.74 & 32.08 & 11.63 & 48.55 & 36.76 \\ 
\multirow{3}{*}{ARC heavy}  & Naive GRPO         & 51.30 & 36.85 & 23.10 & 10.39 & 49.91 & 34.31 \\ 
                             & \ours-Log       & 56.70 & 42.69 & 32.76 & 10.94 & 48.29 & 38.28 \\ 
                             & Dr. GRPO           & 53.70 & 31.50 & 18.70 & 11.08 & 49.57 & 32.91 \\ 
\bottomrule
\end{tabular}
}
\caption{Detailed results showing EM accuracy (\%) on individual datasets and the average, comparing alignment methods across models and domain-heavy training distributions (G=4).}
\label{tab:detailed_results_multirow} 
\end{table}

\begin{table}[h]
\centering
\resizebox{\linewidth}{!}{%
\begin{tabular}{lccccc}
\hline
\textbf{Model} & \textbf{Math-heavy} & \textbf{IMDB-heavy} & \textbf{NQ-heavy} & \textbf{ARC-heavy} & \textbf{Avg.} \\
\hline
\multicolumn{6}{c}{\textit{Qwen2.5 7B}} \\
Naive GRPO & 60.18 & 61.38 & 61.94 & \textbf{60.88} & 61.07 \\
DISCO      & \textbf{60.52} & \textbf{61.74} & \textbf{63.16} & 60.85 & \textbf{61.57} \\
\hline
\multicolumn{6}{c}{\textit{Gemma2-2b-it}} \\
Naive GRPO & \textbf{56.85} & 57.29 & 56.78 & 56.21 & 56.78 \\
DISCO      & 56.68 & \textbf{57.33} & \textbf{56.96} & \textbf{56.51} & \textbf{56.87} \\
\hline
\multicolumn{6}{c}{\textit{Olmo2-1B}} \\
Naive GRPO & 34.78 & 34.57 & 33.38 & 33.97 & 34.18 \\
DISCO      & \textbf{36.61} & \textbf{34.74} & \textbf{34.65} & \textbf{34.74} & \textbf{35.19} \\
\hline
\end{tabular}
}
\caption{Performance comparison of Naive GRPO vs. DISCO across domains.}
\label{tab:additional_result}
\end{table}

\subsection{Reward Curve}
\begin{figure}
    \centering
    \includegraphics[width=\linewidth]{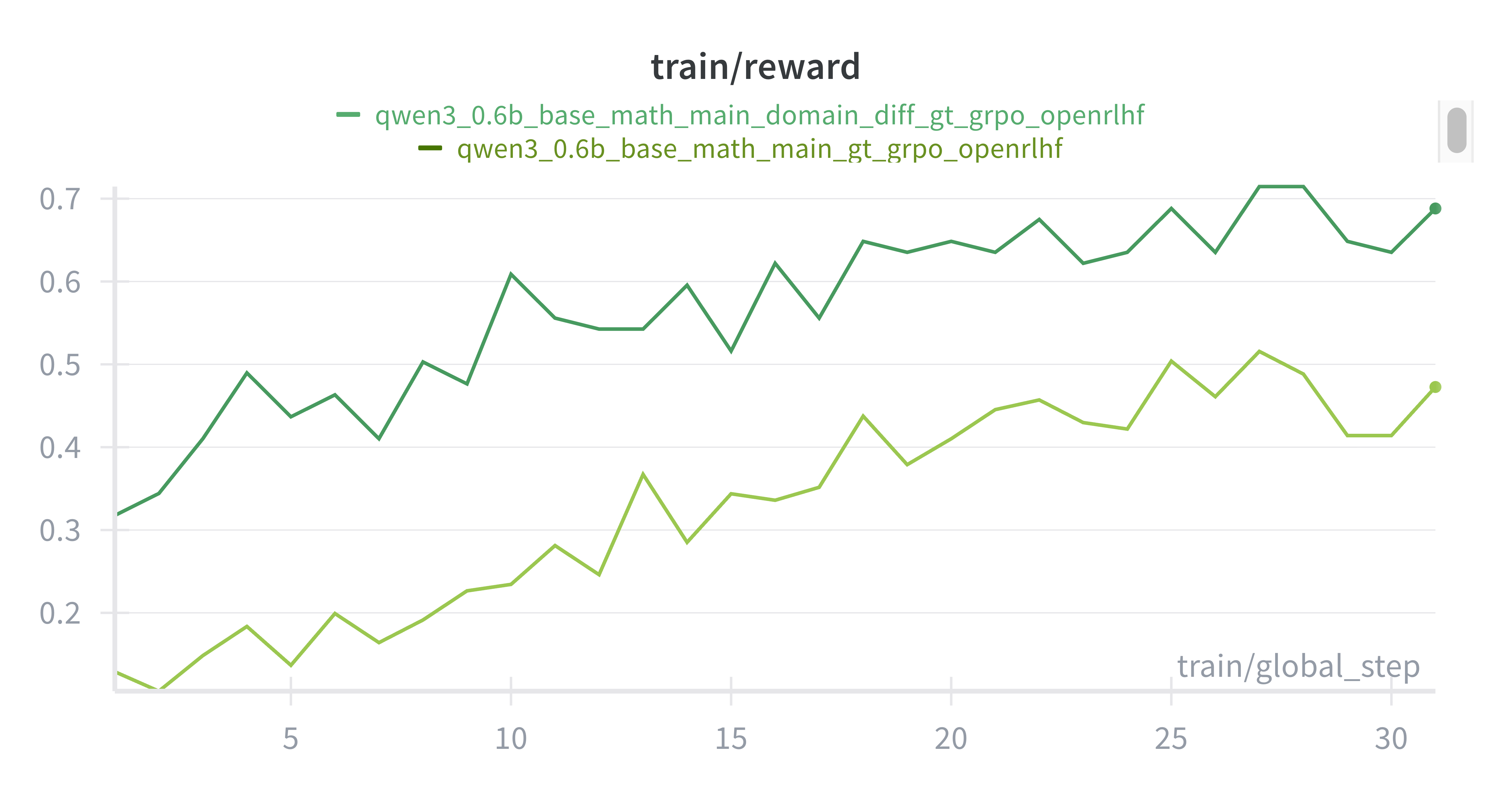}
    \caption{Training reward curves of Qwen3-0.6B models on math heavy training data for Naive GRPO (lighter green line) and \ours (darker green line).}
    \label{fig:reward_curve}
\end{figure}

To better understand how each component contributes during training, we visualize the reward trajectories of Naive GRPO and \ours on Qwen3-0.6B in Figure~\ref{fig:reward_curve}.

\end{document}